\documentclass[letterpaper, 10 pt, conference]{ieeeconf}  

\IEEEoverridecommandlockouts                              

\usepackage{graphicx}
\usepackage{color,soul}
\usepackage{dblfloatfix}
\usepackage{multirow}
\usepackage{multicol}
\usepackage{hyperref}
\usepackage{xcolor}
\usepackage{graphicx}
\usepackage{svg}
\usepackage{amsmath}
\usepackage{amsfonts}
\usepackage{amssymb}
\usepackage{array}
\usepackage{url}
\usepackage[noadjust]{cite}
\usepackage{siunitx}
\usepackage{mathabx}

\begin{document}
\title{\LARGE \bf 
Using Augmented Reality to Assess and Modify Mobile \\ Manipulator Surface Repair Plans}

\author{\protect\parbox{\textwidth}{\protect\centering Frank Regal, Steven Swanbeck, Fabian Parra, Jared Rosenbaum, and Mitch Pryor}
\thanks{All authors are with the Nuclear and Applied Robotics Group (NRG) at The University of Texas at Austin. Austin, TX, 78712, USA. {\tt \{fregal, stevenswanbeck, fabianparra, jaredsr2, mpryor\}@utexas.edu}\vspace{2ex}}
}
\maketitle
\begin{abstract}
Industrial robotics are redefining inspection and maintenance routines across multiple sectors, enhancing safety, efficiency, and environmental sustainability. In outdoor industrial facilities, it is crucial to inspect and repair complex surfaces affected by corrosion. To address this challenge, mobile manipulators have been developed to navigate these facilities, identify corroded areas, and apply protective coatings. However, given that this technology is still in its infancy and the consequences of improperly coating essential equipment can be significant, human oversight is necessary to review the robot's corrosion identification and repair plan. We present a practical and scalable Augmented Reality (AR)-based system designed to empower non-experts to visualize, modify, and approve robot-generated surface corrosion repair plans in real-time. Built upon an AR-based human-robot interaction framework, Augmented Robot Environment (AugRE), we developed a comprehensive AR application module called Situational Task Accept and Repair (STAR). STAR allows users to examine identified corrosion images, point cloud data, and robot navigation objectives overlaid on the physical environment within these industrial environments. Users are able to additionally make adjustments to the robot repair plan in real-time using interactive holographic volumes, excluding critical nearby equipment that might be at risk of coating overspray. We demonstrate the entire system using a Microsoft HoloLens 2 and a dual-arm mobile manipulator. Our future research will focus on evaluating user experience, system robustness, and real-world validation.
\end{abstract}

\section{INTRODUCTION}
\label{sec:introduction}
Industrial robotics are now utilized for routine inspection and maintenance tasks in the nuclear, oil \& gas, and chemical industries \cite{shuklaApplicationRoboticsOnshore2016}. Using novel robotic systems, facilities have increased operator safety and efficiency while reducing environmental impacts \cite{shuklaApplicationRoboticsOnshore2016, IncreasingWorkplaceSafety}. In highly corrosive environments, such as outdoor and offshore manufacturing facilities, operators regularly inspect and repair surface corrosion found frequently on critical equipment. Recent developments within our group have enabled mobile manipulators to perform these repetitive surface corrosion inspection and repair tasks. A mobile manipulator can navigate through an environment, accurately identify, and physically repair geometrically complex corroded surfaces by autonomously applying a protective spray coating. Corrosion often encroaches on valves, gauges, and other functional hardware that would be adversely affected if incidentally coated due to inaccurate predictions or overspray. Thus, operators must visualize and verify the identified corrosion and exclude (or provide a buffer around) such equipment in the repair plan prior to the robot performing repairs. An intuitive user interface to review, modify, and approve the robot's repair plan before it can perform work is needed  to ensure repairs are performed properly. To meet this need, we present an Augmented Reality (AR) Head Mounted Display (HMD)-based system that enables users to a) visualize the point cloud of the identified corroded area, b) define excluded areas the robot should avoid in its repairs, and c) send approval to the robot to proceed with the reviewed repair plan.


\begin{figure}[t!]
    \vspace*{1.2ex}
    \centering
    \includegraphics[width=\linewidth]{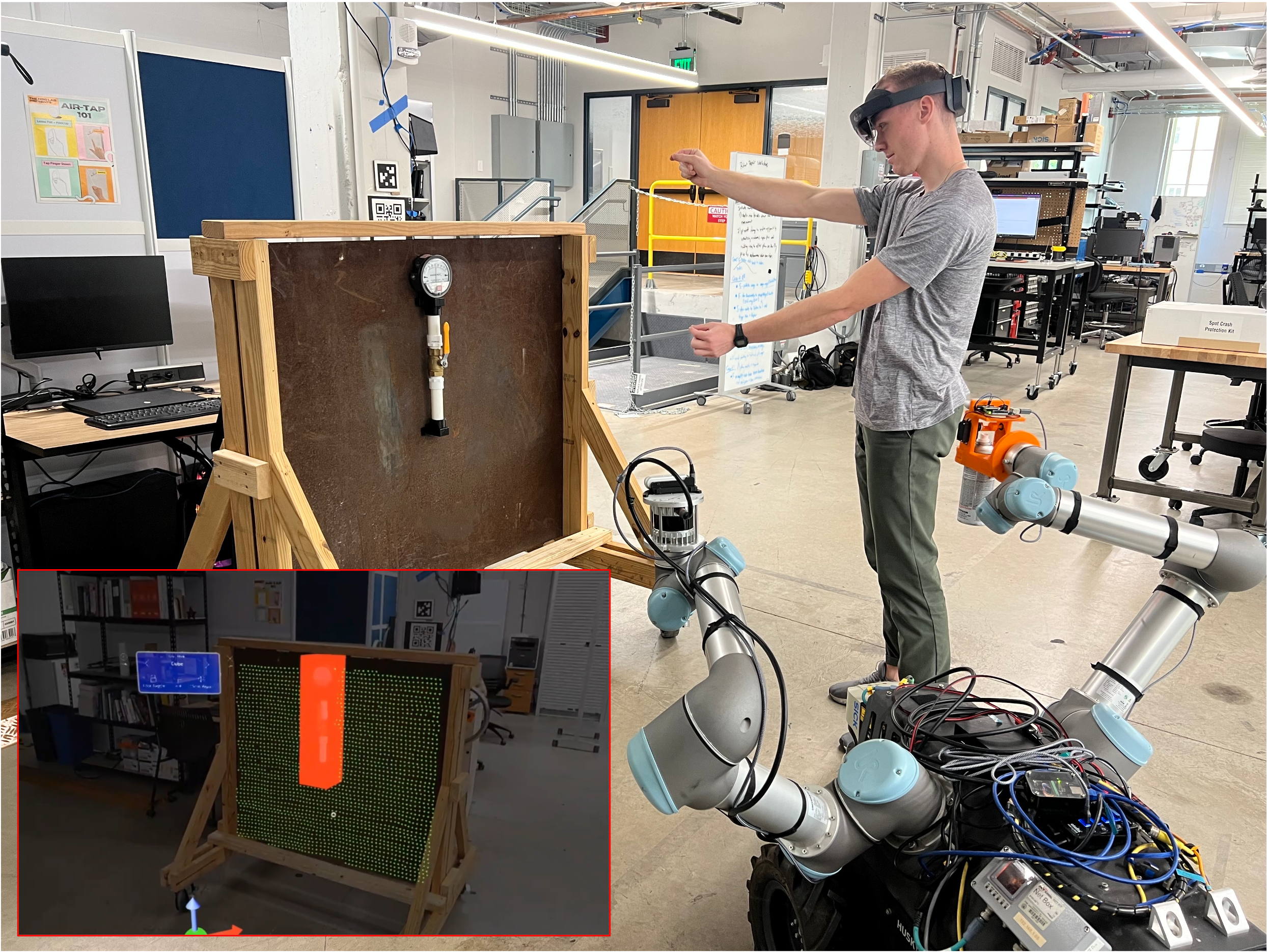}
    \caption{Operator visualizing and modifying point cloud data with a Microsoft HoloLens 2 (FPV, bottom left) from a mobile manipulator (bottom right) that previously identified and now wants to coat the corroded surface. The user is manipulating the red holographic \emph{exclusion volume} (bottom left) to occlude and add a buffer around the value and pressure gauge affixed to the corroded plate.}
    \label{fig:robot-repair-main}
    \vspace*{-4ex}
\end{figure}

The new capability is demonstrated\footnote{Demonstration Video: \url{https://utnuclearroboticspublic.github.io/ar-star/}} on a dual-arm mobile manipulator which autonomously identifies and repairs (or temporarily abates) surface corrosion. The mobile manipulator uses both RGB-D and LiDAR sensors to detect and localize corroded material. 



\begin{figure*}[t!]
    \vspace*{1.2ex}
    \centering
    \includegraphics[width=\linewidth]{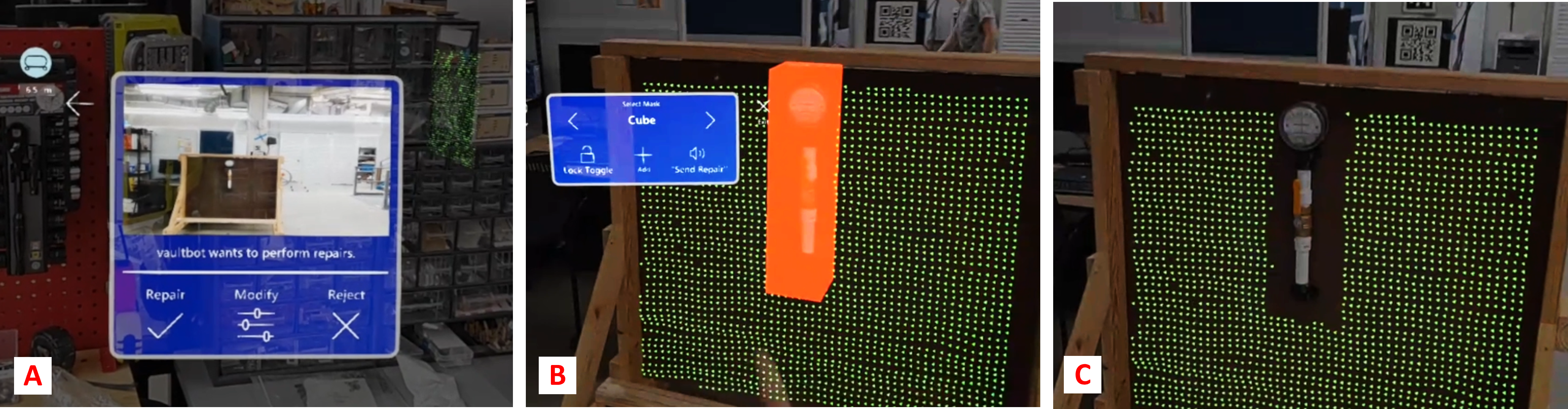}
    \caption{[A] The \textit{Detected Item} menu appears for the AR-HMD user when the robot makes a new positive detection. The user can proceed with the repair, make modifications to the repair plan, or reject the plan. [B] Point cloud of the identified corroded surface prior to modifications; red \emph{exclusion volume} placed on the surface. \textit{Modification} menu in the top left. [C] Modified point cloud after \emph{exclusion volume} was placed to avoid the sensitive equipment on the surface.}
    \label{fig:tri-fecta}
    \vspace*{-2ex}
\end{figure*}
A custom electromechanical device, attached to a 6DoF arm, is used to apply a protective coating to these surfaces (Fig. \ref{fig:robot-repair-main}). Currently, a 2D ROS-based RVIZ interface is used to accept, modify, or reject the surface repair plan. To deploy in industrial settings, users must familiarize themselves with ROS and RVIZ and train extensively with the 2D interface to develop spatial acuity. Research conducted by \cite{rosen_communicating_2017} compared the effectiveness of AR-HMDs to traditional 2D displays. In their study, users were asked to perform a task related to determining robot collisions with both interfaces. The findings revealed that participants using the AR-HMD exhibited a \SI{16}{\%} increase in accuracy and completed tasks \SI{62}{\%} faster than those using the 2D display. These results suggest that by adopting AR-HMDs for this human-robot interaction (HRI), users can significantly enhance the efficiency, awareness, and overall user experience when working with the surface repair robot. Therefore, we leverage our prior work, AugRE \cite{regalAugre2022}, and develop an AR-HMD-based interface to enable users to accept, modify, or reject the surface repair plan.

To localize users and manipulators to a common Cartesian reference frame, we use Azure Spatial Anchors (ASAs). Additionally, we leverage AugRE's Robofleet \cite{sikand_robofleet_2021} based bilateral communication pipeline to transfer data between a Microsoft HoloLens 2 AR-HMD and the ROS-based mobile manipulator. With a localization and communication framework in place, ROS-based nodes on the robotic manipulator were developed to transmit image, point cloud, and projected goal pose data to the AR-HMD user. An AR application module - Situational Task Accept and Repair (STAR) - was added to AugRE. The STAR module enables users to visualize the scene images, the relevant surface point cloud, and a robot goal pose superimposed over the physical world. Users can accept, reject, or exclude portions of the corroded surface from the robot repair plan through this module. The following section summarizes the system components in more detail.

\section{SYSTEM OVERVIEW} 
\label{sec:system-overview}
To enable the STAR HRI, software packages used in parallel on the mobile manipulator and the AR-HMD were developed. This framework takes data processed onboard from the mobile manipulator, and communicates it to a Microsoft HoloLens 2. The STAR module enables visualization and interaction with the data received from the robot's sensors. 

The mobile manipulator system uses input streams of RGB images and 3D LiDAR data to search for corroded material. These data streams are fused and The University of Texas at Austin-developed detection models are used to make predictions about the locations of corroded material in the environment. The geometries of the detected corroded surfaces are extracted and split into actionable, spatially-distinct clusters to be viewed by the user. The locations and geometries of these point cloud clusters are used to generate the repair plans for the robot that fully cover each detected surface. A target goal pose for the robot base within the environment is also calculated using considerations of the robot dimensions, manipulability, and geometry for each surface, such that when placed at the goal pose, the robot will be able to perform the full repair on the target cluster. Once a surface is identified with possible corroded material, the AR-HMD user receives the original image of the scene, point cloud data associated with the identified surface, and a goal pose via ROS messages. 


When an image of the corroded equipment is detected, users are prompted with a notification icon in their scene. This notification icon is clickable using built-in Microsoft HoloLens 2 gesture air-taps. A scrollable list of all detected items is presented when the notification is opened. By tapping an item in this list, a holographic point cloud is aligned and superimposed over the corroded surface, a robot base goal pose is spawned to show where the robot base must be positioned to perform repairs, and a holographic \emph{Detected Item} menu is presented to the user. This menu contains an image of the corroded surface and three holographic buttons (Fig. \ref{fig:tri-fecta}-A). These clickable AR buttons allow the user to command the robot to ``Repair'' the surface, ``Modify'' the observed corrosion, or ``Reject'' the repair altogether. A ROS message is returned to the mobile manipulator on each button press to inform the robot of the user's decision. For ``Repair'' or ``Reject'', the robot makes the coating repairs autonomously or disregards the repair and continues to search for more corrosion. ``Modify'' prompts the user with a \textit{Modification} menu (shown in Fig. \ref{fig:tri-fecta}-B). This menu lets users add various \emph{exclusion volume} shapes (\emph{i.e.} rectangular prism, cylinder, sphere, etc.) to their environment (Fig. \ref{fig:tri-fecta}-B). These shapes are manipulable via pinch-and-hold gestures, and can be dragged and dropped over the point cloud in the scene to designate areas which should be removed from the overall repair plan. When a user clicks or voices ``Send Repair'', a custom message filled with an enumerated shape type, relevant dimensions, and pose of each \emph{exclusion volume} in the scene is sent to the robot. The pose and dimensions of each holographic \emph{exclusion volume} placed by the AR-HMD users are used to extract the inlier points in the original point cloud (Fig. \ref{fig:tri-fecta}-C). Using the modified point cloud, a revised repair plan is regenerated by calculating a series of virtual fixtures normal to and offset by a specified distance from each location on the surface at a target density. The end-effector of the robot is then planned to achieve these virtual fixture poses successively to assess its performance on the surface coverage task, re-planning when a calculated virtual fixture pose is unattainable until a plan for maximum coverage of the entire surface is created. Once the new plan is generated, the updated point cloud is shared again with the user.

After the robot is authorized to execute the repair task from the AR-HMD user, the robot navigates to the calculated target goal pose. Depending on the desired level of autonomy, this can be done by sending the robot base goal pose to the ROS navigation stack, which plans a path to the pose using Dijkstra's algorithm \cite{dijkstra_1959}, or by tele-operating the robot into position, which is aided by visualizing the goal pose in the AR-HMD to ensure no obstructions are blocking the goal pose and ensure the robot is positioned correctly to maximize the success of the repair. Following navigation to the desired location, the robot proceeds with surface coverage repairs.


\begin{figure}[t!]
    \vspace*{1.2ex}
    \centering
    \includegraphics[width=\linewidth]{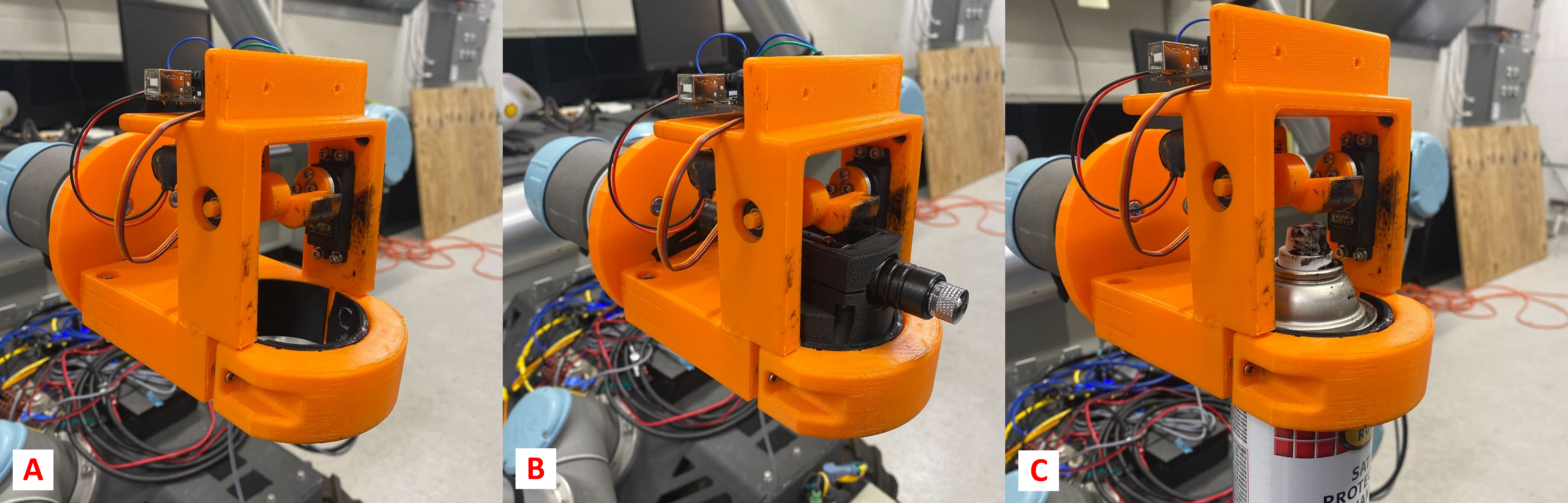}
    \caption{Spray mechanism variations. [A] Mechanism with nothing attached. [B] Mechanism with green laser used for non-destructive coverage tests. [C] Mechanism equipped with Rust-Oleum aerosol can.}
    \label{fig:spray-mechanism}
    \vspace*{-3ex}
\end{figure}


\begin{figure}[t!]
    \vspace*{1.2ex}
    \centering
    \includegraphics[width=\linewidth]{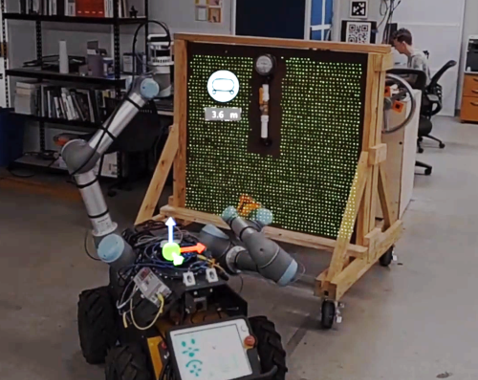}
    \caption{Surface repair robot and corroded iron plate with sensitive equipment attached. The robot is positioned in the environment aligned to the holographic coordinate frame viewed from a Microsoft HoloLens 2. The robot determined this goal pose as the optimal position to complete the surface coverage task.}
    \label{fig:robot_at_goal_pose}
    \vspace*{-3ex}
\end{figure}

\section{TEST IMPLEMENTATION} 
\label{sec:applications}


We demonstrate this work on hardware in an industrial-like indoor lab setting (see video linked in Section \ref{sec:demo-vid}). We used a Microsoft HoloLens 2 AR-HMD and a dual-arm mobile manipulator composed of a Clearpath Husky mobile base with two attached 6DoF Universal Robots UR5 manipulators (Fig. \ref{fig:robot-repair-main} \& \ref{fig:robot_at_goal_pose}) as the surface repair robot. The left manipulator was equipped with an Ouster OS1-128 LiDAR sensor and an Intel RealSense D435i RGB-D camera for corrosion detection. The right manipulator carried a custom electro-mechanical device used to hold and spray protective coatings from aerosol paint cans (Fig. \ref{fig:spray-mechanism}-C). Due to the destructive nature of physically coating test surfaces, the manipulator was not equipped with an aerosol paint can for this demonstration, but the spray trigger was active throughout and future studies will use a laser (as shown in Fig. \ref{fig:spray-mechanism}-B) to mimic permanent surface coatings. A corroded iron plate was used as a sample test surface (Fig. \ref{fig:robot_at_goal_pose}). A $1" \diameter $ PVC pipe segment, with an attached industry standard ball valve and pressure gauge, was affixed to the iron plate and designated as sensitive equipment. With the iron plate mounted on wheels and the sensitive equipment magnetically mounted to the plate, we can easily relocate the test surface (indoors and outdoors) and reconfigure our test environment for future studies.

During the demonstration, an AR-HMD user and a mobile manipulator were assigned specific tasks. Initially, the AR-HMD user's task was to organize a workstation, while the mobile manipulator's task was to survey its environment for corroded surfaces. As autonomous navigation wasn't the primary focus, the manipulator was tele-operated, from an external participant to a position where the test surface was clearly visible to the detection sensors. Once in view, the detection of the corroded test surface was instantiated. An RGB image capturing the test surface, along with a point cloud of the detected corrosion and a goal pose, was sent to the AR-HMD user. Upon receiving a notification in the AR-HMD, the user clicks on the notification to view the image, point cloud, and goal pose presented in the environment. When the user observed the image of the test surface along with sensitive test equipment (a PVC pipe segment), they clicked on the "Modify" button in the \emph{Detected Item} menu. The user then approached the test surface, and upon closer examination of the green point cloud overlaying the iron plate, the AR-HMD user realized that the manipulator's planned spraying trajectory, aimed at covering the entire point cloud, would result in spraying too close to the sensitive equipment (as depicted in Fig. \ref{fig:tri-fecta}-B). To address this issue, the user adjusted the default red holographic rectangular prism, referred to as the \emph{exclusion volume}. They scaled, translated, and rotated this volume to mask the sensitive equipment and also included a buffer zone to prevent overspray. Once satisfied with the placement of the \emph{exclusion volume}, the user selected "Send Repair." Subsequently, the manipulator processed the information regarding the \emph{exclusion volume}, updated the original surface point cloud accordingly, and sent the revised cloud back to the AR-HMD user. The user confirmed the modifications to the plan after visualizing them again. The robot was then tele-operated to reach its intended goal position (as shown in Fig. \ref{fig:robot_at_goal_pose}), planned its trajectory for covering the entire point cloud, and proceeded with executing the plan, activating the spray mechanism as it moved.

\section{CONCLUSION} 
\label{sec:results}
In this work, we present an AR-HMD system that enables users to quickly visualize and modify autonomously detected corrosion \textit{in situ}. The AR-HMD interface allows the robot's intent to be clearly understood and assessed prior to the execution of a repair task. Using images of the scene and a point cloud of the detected corroded surface, the user can assess what the robot intends to repair and permit repair of accurately-detected material, reject problematic task plans for any reason, and make modifications to repair plans to avoid sensitive equipment proximal to corroded material. The user can use other existing functionality \cite{sharp_semiautonomous_2017, regalAugre2022, AugmentedRobotEnvironment2023} to guide the robot into position using the visualized marker of the robot's desired goal pose, aiding navigation through the spatially complex environments intended for deployment. Once a satisfactory repair plan has been created and approved via the AR-HMD user interface, the robot is navigated into position where it then is able to perform its surface coverage tasks. This works shows promise to improve user situational awareness and the overall human-robot interaction when working with mobile manipulators tasked with performing surface repairs. Further research intends to study user trust and performance using this AR-HMD-based interaction system as well as integrate needed functionality from ongoing field tests. 

\section{DEMONSTRATION VIDEO}
\label{sec:demo-vid}
\centering \url{https://utnuclearroboticspublic.github.io/ar-star/}


\typeout{}
\bibliographystyle{IEEEtran}
\bibliography{IEEEabrv,bibliography}



\end{document}